\documentclass[conference]{IEEEtran}

\makeatletter

\def\ps@IEEEtitlepagestyle{%
  \def\@oddfoot{\mycopyrightnotice}%
  \def\@evenfoot{}%
}
\def\mycopyrightnotice{%
  {\footnotesize XXX-X-XXXX-XXXX-X/XX/\$XX.00~\copyright~20XX IEEE\hfill}
  \gdef\mycopyrightnotice{}
}

\usepackage{blindtext}
\usepackage{eso-pic}
\IEEEoverridecommandlockouts
\usepackage{cite}
\usepackage{amsmath,amssymb,amsfonts}
\usepackage{algorithmic}
\usepackage{graphicx}
\usepackage{textcomp}
\usepackage{xcolor}
\def\BibTeX{{\rm B\kern-.05em{\sc i\kern-.025em b}\kern-.08em
    T\kern-.1667em\lower.7ex\hbox{E}\kern-.125emX}}
    
\usepackage{eso-pic}
\newcommand\AtPageUpperMyright[1]{\AtPageUpperLeft{%
 \put(\LenToUnit{0.17\paperwidth},\LenToUnit{-2cm}){%
     \parbox{0.9\textwidth}{\raggedleft\fontsize{8}{11}\selectfont #1}}%
 }}%
\newcommand{\conf}[1]{%
\AddToShipoutPictureBG*{%
\AtPageUpperMyright{#1}
}
} 
\usepackage{bm}
\newcommand{\norm}[1]{\lVert #1 \rVert }
\newcommand{\abs}[1]{\lvert #1 \rvert }
\newcommand{\qq}[1]{\quad \text{#1}\quad}

\newcommand{\tp}{\mathsf{T}}
\newcommand{\dist}[1]{\bm{d}^{\mathrm{#1}} }
\newcommand{\code}[1]{\texttt{#1}}

\DeclareMathOperator*{\argmax}{arg\,max}
\DeclareMathOperator{\sign}{sign}

\begin{document}
\title{\vspace*{1cm} Recognition of Geometrical Shapes \\by Dictionary Learning
}

\author{\IEEEauthorblockN{1\textsuperscript{st} Alexander K\"ohler}
	\IEEEauthorblockA{\textit{Chair of Applied Mathematics} \\
		\textit{Brandenburg University of Technology (BTU)}\\
		Cottbus, Germany \\
		koehlale@b-tu.de}
\and
\IEEEauthorblockN{2\textsuperscript{nd} Michael Breuß}
\IEEEauthorblockA{\textit{Chair of Applied Mathematics} \\
		\textit{Brandenburg University of Technology (BTU)}\\
		Cottbus, Germany \\
		breuss@b-tu.de}
}

\maketitle
\conf{\textit{  Proc. of International Conference on Artificial Intelligence, Computer, Data Sciences and Applications (ACDSA 2025) \\
		7-9 August 2025, Antalya-Türkiye}}
\begin{abstract}
	Dictionary learning is a versatile method to produce an overcomplete set of vectors, called atoms, to represent a given input with only a few atoms. In the literature, it has been used primarily for tasks that explore its powerful representation capabilities, such as for image reconstruction. In this work, we present a first approach to make dictionary learning work for shape recognition, considering specifically geometrical shapes. As we demonstrate, the choice of the underlying optimization method has a significant impact on recognition quality. Experimental results confirm that dictionary learning may be an interesting method for shape recognition tasks.
\end{abstract}



\section{Introduction}


Dictionary Learning or Sparse Dictionary Learning (SDL) was proposed by Olshausen and Field \cite{Olshausen1996}, motivated by their study of the receptive fields of simple cells in mammalian primary visual cortex.
The method aims to find a sparse representation of the input data by a linear combination of so-called atoms.
The collection of these atoms is called a dictionary. One unique selling point of this method is, that these atoms do not need to be orthogonal to each other.
Therefore, in general, the number of atoms in a dictionary exceeds the dimensionality of the input set, i.e., it is overcomplete. 
As shown e.g., in \cite{Elad2010,zhang2022dictionary}, the representation capability 
of dictionary learning has proven useful for image processing
applications like e.g., 
denoising~\cite{giryes2014sparsity},
deblurring~\cite{ma2013dictionary,xiang2015image}, or
image enhancement~\cite{cao2014segmentation}.

Solving the SDL problem for the input data $Y$ means finding a dictionary $D$ and a representation matrix $X$, so that $Y \approx DX$.
Both matrices $D$ and $X$ are unknown and need to be determined via 
the solution of an optimization problem.
An established method to solve 
$Y \approx DX$ 
under sparsity constraints is to formulate it in terms of two stages,
and solve them alternately one after the other. 
This idea is commonly employed, like e.g., in the presentation of the 
K-SVD algorithm by Aharon~\cite{Aharon2006}. Let us note that usually
the second stage in this setting tackles the sparsity problem, and 
it is in general NP-hard~\cite{Foucart2013} so that in practice, often
a relaxation of the problem is employed at some stage.

There are multiple algorithms to approximately solve the sparsity problem.
The methods of interest here are the Orthogonal Matching Pursuit (OMP) method and the Least Angle Regression Stagewise (LARS) method.
The OMP method was introduced by Mallat in 1993~\cite{Mallat1993} to decompose any signal into a linear expansion of waveforms that are selected from a redundant dictionary of functions.
The LARS method was presented in 2004 by Efron et al.~\cite{Efron2004} to combine the benefits of forward selection algorithms and backward elimination.


To our best knowledge, dictionary learning has not been used before
for the shape recognition task.
Within the first approach we present in this work, 
the underlying feature vector generation 
is supposed to be a delicate task. Therefore, we discuss this point in detail,
to enable a more profound understanding of important aspects 
of dictionary learning itself and how it acts in a shape recognition setting.
Specifically, we show here that the choice of the optimization method 
has a significant influence on the quality of classification results.

\section{Mathematical Methods and Notations\label{sec:math}}

After reviewing SDL and the
OMP/LARS methods, we introduce the
notation for shapes and their classes.

\subsection{Dictionary Learning\label{sec:dic_learn}}
We start with an explanation of the SDL procedure.
As indicated, it allows us to construct a non-orthogonal and 
overcomplete representation called a \emph{dictionary}.
In the context of this paper, the learned dictionary is a matrix $D\in \mathbb{R}^{N \times K}$, with $K\gg N$.
This property ensures the overcompleteness.
The columns of this dictionary, $\lbrace d_i \rbrace_{i=1}^K$, $d_i \in \mathbb{R}^N$, will be called \emph{atoms}.

To learn the dictionary $D$, we need to provide a training dataset $\lbrace y_i \rbrace_{i = 1}^M$ containing $M\in \mathbb{N}$ vectors $y_i \in \mathbb{R}^N$.
Analogously to the dictionary, we can represent the dataset via a matrix $Y \in \mathbb{R}^{N\times M}$.

Learning the dictionary $D$ from the dataset $\lbrace y_i \rbrace_{i = 1}^M$ is synonymous with solving the optimization problem
\begin{equation}
	\label{eq:dict_learning}
	\min_{D,x_i} \sum_{i = 1}^{M} \norm{y_i - Dx_i}_2^2 \qq{w.r.t.} \norm{x_i}_0 \leq T
\end{equation}
with the representation vectors $x_i \in \mathbb{R}^K$, $i\in \lbrace 1,\ldots, M\rbrace$, and the parameter $T\in \mathbb{N}$.
With $\norm{\cdot}_0$, we declare the \emph{zero norm}, which counts the non-zero elements of a given vector and $\norm{\cdot}_2$ denotes the Euclidean norm.
The vectors $x_i$ will act as an activation for the different atoms within the matrix $D$.
Now, the dataset $Y$ can be represented via $Y \approx DX$, where $X=\lbrace x_i\rbrace_{i=1}^{M} \in \mathbb{R}^{K \times M}$ is 
the \emph{representation matrix}.

As indicated, solving the optimization problem 
\eqref{eq:dict_learning} is NP-hard~\cite[chapter 2.3]{Foucart2013}.
A common approach to dealing with this issue, is to solve the problem in two steps.
These steps are performed subsequently until the conditions in \eqref{eq:dict_learning} are met.

The first step consists of the problem
\begin{equation}
	\label{eq:dict_learning_subproblem_1}
	\min_{D} \sum_{i = 1}^{M} \norm{y_i - Dx_i}_2^2
\end{equation}
and the second step solves the problem
\begin{equation}
	\label{eq:dict_learning_subproblem_2}
	\min_{x_i} \sum_{i = 1}^{M} \norm{y_i - Dx_i}_2^2 \qq{w.r.t.} \norm{x_i}_0 \leq T
\end{equation}
The first problem \eqref{eq:dict_learning_subproblem_1} is comparatively 
simple to solve.
There are several algorithms to solve the significantly harder problem \eqref{eq:dict_learning_subproblem_2}.
Two of them, we briefly explain in the next sections.


\subsection{Orthogonal Matching Pursuit}

In this section, we briefly review the orthogonal matching pursuit (OMP) algorithm. For more details, we refer to \cite[Chapter~3.1.2]{Elad2010} or \cite[Chapter~3.2]{Foucart2013}.

The OMP algorithm belongs to the family of greedy methods and 
solves the optimization problem
\begin{equation}
	\label{eq:p0}
	(P_0): \qquad \min_{\bm{x}} \norm{\bm{x}}_0 \quad \text{subject to}\quad \bm{b} = A \bm{x}
\end{equation}
Where $A$ is the matrix $A = \left[\bm{a}_1, \bm{a}_2, \ldots, \bm{a}_K\right] \in \mathbb{R}^{n \times K}$, the vector $\bm{a}_i$ denotes $i$th column of the matrix $A$, and the $\bm{b}\in \mathbb{R}^n$ are the constant terms of the linear equation.

OMP solves this problem iteratively. The process starts with
setting the initial solution $\bm{x}^0 = \bm{0}$, the initial residual $\bm{r}^0 = \bm{b}-A\bm{x}^0 = \bm{b}$, and the initial support $S^0 = \emptyset$. In each step, OMP computes the errors
\begin{equation}
	e(j) = \min_{\alpha_j} \norm{\alpha_j\bm{a}_j  - \bm{r}^{k-1} }_2^2
\end{equation}
for all $j\in \lbrace 1, \ldots, K \rbrace$, 
with an optimality factor $\alpha^* = \frac{\bm{a}_j^\tp \bm{r}^{k-1} }{\norm{\bm{a}_j}_2^2}$.
After computing all errors $e(j)$, OMP adds the index $j$ that gives the smallest error and is not already part of the support $S^{k-1}$ to the support $S^k = S^{k-1} \cup \{j\}$.

After updating the support,
OMP will find $\bm{x}^k$ by the minimization of $\norm{A_{S^k} \bm{x}_{S^k} - \bm{b} }_2^2$,
with $A_{S^k} \in \mathbb{R}^{n \times \abs{S^k}}$ being
the matrix $A$ reduced to the columns that are in the support $S^k$,
and $\bm{x}_{S^k}$ the non-zero elements of the vector $\bm{x}^k$.

The last step, within the iteration $k$, would be updating the residual $\bm{r}^k = \bm{b} - A\bm{x}^k$.
These steps will be repeated until $\norm{\bm{r}^k}_2 < T$ for some threshold $T\in \mathbb{N}$.

In each iteration, OMP chooses a solution that is best in this iteration, this is why it is part of the greedy method family.
The orthogonality defining OMP stems from the fact, that 
chosen columns of $A$ that are in the support are orthogonal 
to the residuals.

\subsection{Least Angle Regression Stagewise}
The main idea of least angle regression stagewise (LARS) has been
introduced in \cite{Efron2004}. We will provide only the basics to this algorithm.
For more in-depth expositions, we refer to the 
original article \cite{Efron2004} or \cite[Chapter~5.3.3]{Elad2010}.

LARS is designed to solve the optimization problem
\begin{equation}
	\label{eq:p0epsilon}
	(P_0^\epsilon): \qquad \min_{\bm{x}} \norm{\bm{x}}_0
	\qq{subject to}
	\norm{\bm{b} - A \bm{x}}_2 \leq \epsilon
\end{equation}
To this end, LARS builds up estimates $\bm{\mu} = A\bm{x}$.
With each step in the algorithm, LARS is updating $\bm{x}$ so that after $k$ steps, only $k$ elements of $\bm{x}$ are non-zero.

LARS starts with the initialization of an index set $S = \emptyset$ and $\bm{\mu}_{S} = \bm{0}$.
Furthermore, it expects that $\bm{b}$ and the columns of $A$ are centered, and that the columns of $A$ are normalized
\begin{equation}
	\sum_{i = 1}^{n} b_i = 0,~ \sum_{i = 1}^{n} a_{ij} = 0, ~ \text{and} ~ \sum_{i = 1}^{n} a_{ij}^2 = 1
\end{equation}
for $j \in \lbrace 1, \ldots, K \rbrace$.

Within each step of the algorithm,
LARS computes the correlation $\bm{c}$ between the columns of $A$ and a residual vector $\bm{r}_S = \bm{b} - \bm{\mu}_{S}$ via $\bm{c} = A^\tp \bm{r}_S$. The indices with the largest absolute value of $\bm{c}$ are added to the index set $S = S \cup \lbrace i \colon \abs{c_i} = C \rbrace$, with $C = \max_{i}\abs{c_i}$. With the updated index set, we can compute $A_S = \left\lbrace s_i\bm{a}_i \right\rbrace_{i \in S}$, with $s_i = \sign(c_i)$, which is the matrix containing only of columns of $A$ that are in the index set $S$.

With $A_S$ LARS computes $G_S = A_S^\tp A_S$, $T_S = \left(\bm{1}_S^\tp G_S^{-1} \bm{1}_S\right)^{-\frac{1}{2}}$, $\bm{w}_S = T_S G_S^{-1} \bm{1}_S$ and finally $\bm{u}_S = A_S \bm{w}_S$.
Thereby, it uses $\bm{1}_S$, the vector of ones with the same length as the index set $S$.
With $\bm{u}_S$, the algorithm can update the estimate
$\bm{\mu}_S = \bm{\mu}_S + \alpha \bm{u}_S$
using
$\alpha = \min^+_{i\in S^{\mathrm{c}}} \left\lbrace \frac{C-c_i}{T_S-t_i} , \frac{C+c_i}{T_S+t_i} \right\rbrace$
and
$\bm{t} = A^\tp \bm{u}_S$.
Thereby, $\min^+$ denotes that the minimum is taken over positive values and $S^{\mathrm{c}}$ is the complement of $S$.

The algorithm stops if the index set $S$ is full, i.e. $\abs{S} = K$, or if $\det G_S = 0$.

The factor $\alpha$ ensures that in the next step, the correlation $\bm{c}$ attains its maximum value on at least one element more.
Therefore, the index set $S$ is growing in each step. And if the correlation $\bm{c}$ has the same value for two different indices, one knows that the angle between $\bm{r}_S$ and $A$ is the same for both indices.

\subsection{Shapes and shape classes}
In the context of this paper, a shape $\mathcal{S}$ is the realization of a closed curve in a two-dimensional space, $\mathcal{S}  \subset \mathbb{R}^2$.
The discrete version of a shape is established via a point cloud $P\subset \mathbb{R}^2$ with $\abs{P} = n \in \mathbb{N}$ points $(x_i, y_i) = p_i \in P$ for $i \in  \lbrace 1, 2, \ldots, n \rbrace$.
With $x_i$ and $y_i$, we denote the $x$-- and $y$--coordinate of the $i$th point $p_i$ in the point cloud $P$.
Since, we have multiple point clouds, we want to introduce the point cloud set $\mathcal{P}$ to store them.
With $\abs{\mathcal{P}}$, we got the number of point clouds within the set.

For convenience, a \emph{geometrical shape set} is introduced via
\begin{equation}
	O = \bigcup_{s\in G} O_s
\end{equation}
for an arbitrary index set $G=\lbrace g_1, g_2,...g_m\rbrace$ that stores the identification of the $m$ different geometrical shape classes.
A useful function is $\iota\colon \lbrace 1,\ldots, m\rbrace\to G$, $\iota(i) = g_i$, allowing to refer to the elements of the set $G$.

Writing $\bm{v} \in O_s$ for $s\in G$ means that a vector $\bm{v}$ is part of the shape class $O_s$.

\section{Experiments\label{sec:experiments}}

We first describe the experimental set-up before we finally discuss
experimental results.

\subsection{Dataset}

For the experiments, we use a processed version of the shape dataset \cite{Korchi2020}.
The original dataset consists of $10,000$ pictures
of different geometrical shapes,
namely triangle, square, pentagon up to nonagon, circle,
and a star with a pentagonal hole in the center.
In total, this dataset hosts nine shape classes.
These pictures vary in colors (the color of the shape and the color of the background), size, rotation, and position of the shape within the picture.
The first step of the data processing step provides point clouds of the shapes within the images.
We do not elaborate on this step here, details of this process can 
be found in \cite{Koehler2022,Koehler2024}.

Since, the star shape is omitted during the construction of the point clouds,
the dataset of point clouds hosts only eight different shape classes,
$G = \left\{ 3,4,5,6,7,8,9,\infty \right\}$, and $\abs{G} = 8$.
The shape classes are denoted regarding the number of edges of the specific shape class.
Since the circle has no defined number of edges, the symbol $\infty$ was chosen to represent the circle shape class.

It needs to be emphasized that the points within the point cloud are ordered.
This means that the successor $p_{i+1}$ of the point $p_i$, for $p_i, p_{i+1}\in P$, are next to each other in the shape as well.
Another characteristic of the dataset is, that the first and last point in the point cloud refer to the same point on the shape.
So that, if the point cloud has $n$ points, there are, in fact, only $n-1$ different points in the point cloud.

Since the geometrical shapes in the images differ in size, the number of points within the point clouds differ, too.
The number of points ranges from $n\approx100$ up to $n\approx500$.
These differences need to be addressed during the preprocessing.

By the set-up, the dataset that is used for calculations inherits 
the number of point clouds from the number of images of the original dataset, meaning $\abs{\mathcal{P}} = 10,000$.
The expression $\mathcal{P}_s \in O_s$ indicates the entirety of point clouds of a certain shape class $O_s$, $s\in G$.

For learning purposes, it is standard to divide the dataset into two disjunct sets.
One for training~$\mathcal{E}$ and one for testing~$\mathcal{T}$, $\mathcal{P} = \mathcal{E}\dot{\cup} \mathcal{T}$.
We will split the dataset with a ratio of $7:3$, this means, $70\%$  the dataset is the training set $\mathcal{E}$ and $30\%$ is the test set $\mathcal{T}$ ($\abs{\mathcal{E}} = 7,000$ and $\abs{\mathcal{T}} = 3,000$).

With an index $s$, the affiliation to a certain shape class $O_s$ is indicated: $\mathcal{E}_s$ and $\mathcal{T}_s $, for $s\in G$.

\subsection{Preprocessing}

In this section, we want to discuss the preprocessing of the point clouds.
To be more precise, we now discuss the process to transform the point clouds into some data structure useful for the recognition task.
But before we begin with the details of the preprocessing,
we would like to define what we want to achieve by it.

The input vectors $y$ for the recreation within the dictionary learning process, need to be:
{\em (i)} equal in length, regardless of the number of points in the corresponding point cloud, and {\em (ii)} each element of the vector, within a shape class,  should store almost similar meaning, especially certain elements of the distance vector are supposed to be
close to a corner of a shape.
We conjecture that this hidden meaning embedded in $y$ is learned with our approach.

\subsubsection{Distance measure}
\begin{figure}
	\centering
	\includegraphics[height = 0.4\columnwidth]{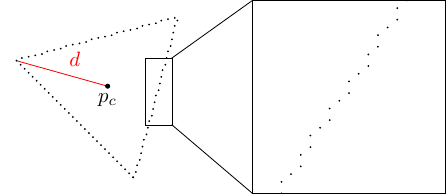}
	\caption{\label{fig:triangle_distance}
		On the left, is the visualization of the distance computation, exampled on a triangle point cloud.
		On the right, an enlargement of the points within the point cloud is shown.}
\end{figure}

\begin{figure*}
	\centering
	\includegraphics[width = \textwidth]{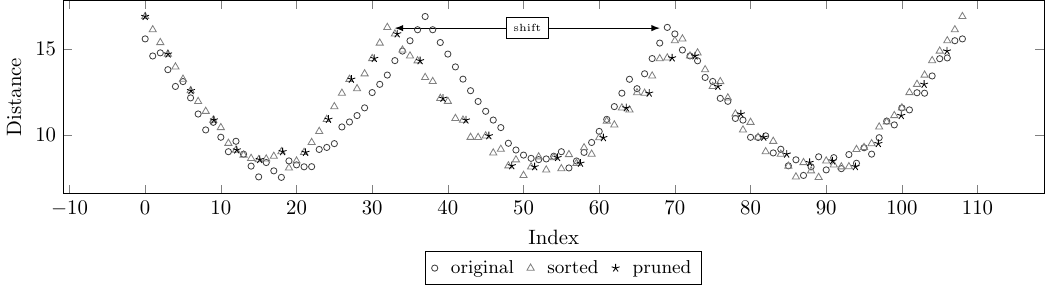}
	\caption{\label{fig:triangle_feature_compare}
		Comparison of the different stages of preprocessing, by the example of a specific triangle shape.
		With the circle marks, we denote the elements of the distance vector $\bm{d}$.
		The triangle marks represent the modified distance vector $\dist{s}$. And $\dist{p}$ is representing the pruned distance vector, indicated with the star marks. Additionally, we indicated the shift between the original and sorted distance vector.
	}
\end{figure*}

The first part in the preprocessing is to construct a {\em distance}, given a point cloud.
The idea explored in this paper is an extension of the D1 distance presented in \cite{Osada2001}.
The D1 distance from that paper describes the distance
$d(\cdot) \colon \mathbb{R} \times \mathbb{R} \to \mathbb{R}$
between a point $p_i \in P$ and the center point $p_c = (x_c,y_c)$ via
\begin{equation}
	d(p_i)  =
	\norm{p_i -p_c}_2
	\quad \text{with} \quad
	p_c = \frac{1}{n-1} \sum_{i = 1}^{n-1} p_i
\end{equation}
Here $\norm{\cdot}_2$ is the Euclidean norm and
the barycenter $p_c$ is computed using the arithmetic average of the $n-1$ unique points of the point cloud, $n = \abs{P}$.
Using all $n$ points would lead to a slight shift towards this doubled point.
The Figure \ref{fig:triangle_distance} (left)
visualizes the points of the point cloud together with the distance from the center point.
On the right, a close-up of the points is shown.

In the end, these distances will be stored in a distance vector $\bm{d}\in\mathbb{R}^n$,
\begin{equation}
	\bm{d} = (d_1, \ldots, d_n)^\tp =\sum_{i=1}^{n} \bm{e}_i d(p_i)
\end{equation}
with $\bm{e}_i$ the $i$th Cartesian unit vector,
and $p_i \in P$ for all $n = \abs{P}$ points from the point cloud $P$.

At this point in preprocessing, the length of different vectors $\bm{d}$
still differs.
Before we address this problem, we want to adjust the distance vectors.

\subsubsection{Normalization}


The fact, that the points are ordered within the point cloud, means that the distance vector $\bm{d}$ is related to a $360^\circ$ scanning of the shape.
Since the point clouds consist of $n$ points, there are $n-1$ increments. 
In the context of the circle, we scan in $\frac{360^\circ}{n-1}$ angle increments. 

From this perspective, it seems natural to set the number of angle increments to a fixed value $N$.
The presented approach uses the distance vector $\bm{d}$ as a source and interpolates, if necessary, needed distances between the elements of $\bm{d}$.

Before we follow this idea, we want to change the ordering within the distance vector. 
Currently, we start randomly somewhere in the shape. 
The idea is to start with a corner of a shape because it is simple to find.
The corners of the considered geometrical objects are the points with the highest distance from the center point.

Therefore, we have to find the index $q$ of the maximum distance $d_q$ within the distance vector $\bm{d}$, $q = \argmax_{i \in \lbrace 1, \ldots ,n \rbrace} d_i$.
With the index~$q$, we can sort the distance vector $\bm{d}$.
We can cut the vector $\bm{d}$ into two parts, 
the part before and including the index $q$, $(d_1, \ldots, d_{q})$,
and the part after and including $q$, $(d_{q}, \ldots, d_{n})$.
Since, $d_1$ and $d_n$ refer to the same point, we can omit one of these two distances.
And with including $d_q$ twice, we keep the length of the vector unchanged.

Now, we reorder the distance vector $\bm{d}$ in the following way to get a \emph{sorted} distance vector $\dist{s}\in \mathbb{R}^n$,
where the first and last element in the vector is always the maximal element:
\begin{equation}
	\dist{s} =
	\left( d_1^{\mathrm{s}}, \ldots , d_n^{\mathrm{s}} \right)^\tp =
	\left(
	d_{q}, \ldots, d_{n-1}, d_1, \ldots, d_{q}
	\right)^\tp
\end{equation}
The length of this sorted distance vector is still inherited 
from the number of points in the point cloud.

After sorting, we scan the sorted distance vector $\dist{s}$ for a fixed number of $N$ equidistant points.
Normally, we want to reduce the length of the distance vector $\dist{s}$, $N < n$, so we will call the new scaled vector \emph{pruned} distance vector, $\dist{p}$.

If the vector $\bm{d}$ consists of $n = 101$ distances, 
then there are $100$ intervals, and the angle of the intervals is $3.6^{\circ}$.

Now we want to use only $N = 10$ distances, $36^{\circ}$ intervals.
Constructing the pruned vector with the help of every $10$th element of the distance vector gives $\dist{p} = (d_1, d_{11}, \ldots, d_{91})^\tp \in \mathbb{R}^N$.
In reality, there is no guarantee that $\frac{n-1}{N} \in \mathbb{N}$.
Hence, we need to include the case that defines what is happening if we need an element \emph{between} the elements of $\dist{s}$.

We construct $\dist{p}\in \mathbb{R}^N$, with the factor $f_i := 1+ (i-1) \cdot \frac{n-1}{N}$, $a_i = \lfloor  f_i \rfloor$, and $b_i = \lceil f_i \rceil$
\begin{equation}
	\dist{p} = \sum_{i = 1}^{N} \bm{e}_i
	\begin{cases}
		d^{\mathrm{s}}_{f_i}                             & \text{if } f_i \in \mathbb{N} \\
		(1-\alpha)d^{\mathrm{s}}_{a_i} + \alpha d^{\mathrm{s}}_{b_i} & \text{if } f_i \in (a_i,b_i)
	\end{cases}
\end{equation}
where $\bm{e}_i$ denotes the $i$th unit vector and $\alpha = b_i - a_i \in (0,1) $.
At this point, we want to emphasize, that during this step ($\dist{s} \to \dist{p}$) we omit the double element in the distance vectors.

A comparison of all three steps of the preprocessing is visualized in Figure \ref{fig:triangle_feature_compare}.
There, the original distance vector $\bm{d}$ is represented with the circle marks. For the sorted distance vector $\dist{s}$ we used the triangle marks and with the start marks we denoted the pruned distance vector $\dist{p}$.
Additionally, the shift between $\bm{d}$ and $\dist{s}$, because of sorting, is highlighted.


\subsection{Dictionary Learning}
All the (labeled for training) pruned distance vectors $\dist{p}_s$ from a certain shape class $O_s$, $s\in G$, form the input data for the dictionary learning process.
\begin{equation}
	Y_s = \lbrace y_i\rbrace_{i=1}^{\abs{\mathcal{E}_s}} = \lbrace d^{\mathrm{s}}_i\rbrace_{i=1}^{\abs{\mathcal{E}_s}} \in \mathbb{R}^{N\times \abs{\mathcal{E}_s}}
\end{equation}
Solving the sparse dictionary learning problem \eqref{eq:dict_learning}, returns the dictionary $D_s\in\mathbb{R}^{N\times K}$ and a representation matrix $X_s \in \mathbb{R}^{K \times \abs{\mathcal{E}_s}}$ for a certain shape class $\mathcal{O}_s$, $s\in G$.

To solve the dictionary learning problem \eqref{eq:dict_learning}, 
we make use of the python library \code{scikit-learn} \cite{scikit-learn}.
For full disclosure, this method minimizes the $\norm{\cdot}_1$ of the representation vector $\bm{x}$, instead of $\norm{\cdot}_0$.
This is necessary because technically the $\norm{\cdot}_0$ is not a real norm, which is crucial for the convergence of algorithms. 
A justification for this norm relaxation can be found in \cite{Candes2005,Ramirez2013}.

This library provides the \code{DictionaryLearning} class.
The relevant settings in the context of this paper are: \\
\code{n\_components} which determines the number of atoms $K$;
\code{transform\_n\_nonzero\_coefs} is an input to control the number of non-zero elements $T$ in the reconstruction vector;
and \code{transform\_algorithm} which allows us to change between the OMP and LARS algorithm.

\subsection{Recognition: set-up and evaluation\label{sec:matching}}
After learning the dictionary $D_s$ and the representation matrix $X_s$ for a certain shape class $O_s$, $s\in G$,
we can move on to decide if $\dist{p}_t$ from an unknown shape class $O_t$, $t\in G$, matches with members of the learned shape class $O_s$.

The approach of the paper is to use the learned representation matrix $X_s = \lbrace x_i \rbrace_{i=1}^{\abs{\mathcal{E}_s}} = \left(\bm{x}_1, \ldots, \bm{x}_{\abs{\mathcal{E}_s}}\right) \in \mathbb{R}^{K \times N}$.
Remember, each vector $\bm{x}$ contains at maximum $T$ non-zero elements, and the index of these non-zero elements tells us which atom from the dictionary $D$ is used.

To make use of this, we construct a \emph{meta-dictionary} by joining the learned dictionaries for the different shape classes $D_s$ together:
\begin{equation}
	D_\mathrm{m} = \left\lbrace D_s \right\rbrace_{s \in G}  = \left(D_3,D_4, \dots, D_{\infty}\right) \in \mathbb{R}^{N\times \abs{G}\cdot K}
\end{equation}
Fixing the dictionary in this way allows us, with the help of \eqref{eq:dict_learning_subproblem_2}, to find the representation vector $\bm{x}^{\mathrm{m}}$ for a distance $\dist{p}$ from an unknown shape class.

The idea is that the atoms, built up from the dictionaries of different shape classes, are distinguished enough,
so that a distance vector of a certain shape class relies more on the atoms from the dictionary of the shape class to which it belongs.

Assuming this is true, checking the indices of the non-zero elements in $\bm{x}^\mathrm{m}$, allows determining which atoms in $D_{\mathrm{m}}$ were used.
The construction of $D_{\mathrm{m}}$, allows drawing conclusions about the shape class by the range of the used indices.
If an index $i\in(0, K]$, it implies that an atom from the first dictionary $D_{3}$ is used, and if $i\in (K, 2K]$, then an atom from the second dictionary $D_4$ is used.
In general, we can say, that if $i\in((l-1)K, lK]$ it is part of the dictionary $D_{\iota(l)}$, using the function $\iota\colon\mathbb{N}\to G$ to index the elements in $G$.

Pouring these thoughts into equations, we will first construct an index vector $\bm{\mathrm{idx}}$, consisting only of the indices of the non-zero elements of the vector $\bm{x}^{\mathrm{m}}$:
\begin{equation}
	\bm{\mathrm{idx}} = \left( i \colon x^{\mathrm{m}}_i \neq 0~ \forall x^{\mathrm{m}}_i \in \bm{x}^{\mathrm{m}} \right)^\tp \in \mathbb{N}^t
\end{equation}
Having $\bm{\mathrm{idx}}$ and dividing it with $K$ as well as rounding the result up to the next integer  will lead to a vector $\bm{m}$ as
\begin{equation}
	\bm{m}
	= \left\lceil\frac{\bm{\mathrm{idx}}}{K}\right\rceil
	= \left( \left\lceil\frac{\mathrm{idx}_1}{K}\right\rceil, \ldots , \left\lceil\frac{\mathrm{idx}_t}{K}\right\rceil \right)^\tp \in \mathbb{N}^t
\end{equation}
This vector stores numbers that are linked, via $\iota(m_i)\in G$, to all shape classes in $O$.
For the length $t$ of the vectors $\bm{idx}$ and $\bm{m}$ applies that $t \leq T$ because it is possible to use less than $T$ atoms for representation.

Counting how often a certain number appears
\begin{equation}
	\mathrm{cnt}(\bm{m}, i) = \frac{1}{t}\sum_{j = 1}^t \delta(m_j, i)
	~\text{with}~
	\delta(v, i) = \begin{cases}
		1 & ,\, v = i \\ 0&,\, v \neq i
	\end{cases}
\end{equation}
leads to a classification vector
\begin{equation}
	\bm{c} = \left(\mathrm{cnt}(\bm{m}, i)\right)_{i=1}^{\abs{G}} \in \mathbb{R}^{\abs{G}}
\end{equation}
This vector stores the percentages of the elements of $m^{\mathrm{c}}$.

A vector $\bm{c} = (0, 0.2, 0.3, 0.5)^\tp$ means that the distance vector $\dist{p}$ is matched with a probability of $20\%$ to shape class $O_{\iota(2)}$,
to $30\%$ to shape class $O_{\iota(3)}$,
and with $50\%$ to the shape class $O_{\iota(4)}$,
assuming in this example that we distinguish only between four shape classes.

Since we know the geometrical shape class $O_s$ of the point clouds within the test set $\mathcal{T}^s$, $s\in G$, this information is passed on to the distance vectors of $\dist{p}_s$.
Consequently, it is justified to write $\bar{\bm{c}}^s$ to denote the 
shape class of the mean classification vector:
\begin{equation}
	\bar{\bm{c}}^s
	= (\bar{c}_1^s,\dots, \bar{c}_{\abs{G}}^s )^\tp
	= \frac{1}{\abs{\mathcal{T}^s}} \sum_{i = 1}^{\abs{\mathcal{T}^s}} \bm{c}^s_i
	\quad \bm{c}^s_i \in \mathcal{T}^s,\, s\in G
\end{equation}

To visualize the recognition results and to evaluate the quality of 
our approach, we introduce a 
{\em hit-rate matrix} $\mathrm{hrm} \in \mathbb{R}^{\abs{G}\times \abs{G}}$
\begin{equation}
	\mathrm{hrm}(s,t) = \lbrace\bar{c}_{\iota(j)}^{s_i} \rbrace_{i,j}^{\abs{G}}
	= \begin{pmatrix}
		\bar{c}_{t_1}^{s_1}         & \dots  & \bar{c}_{t_{\abs{G}}}^{s_1}         \\
		\vdots                           & \ddots & \vdots                                 \\
		\bar{c}_{t_1}^{s_{\abs{G}}} & \dots  & \bar{c}_{t_{\abs{G}}}^{s_{\abs{G}}}
	\end{pmatrix}
\end{equation}
Each row $i$ belongs to an origin shape class according to one $s_i \in G$ and the columns $j$ belong to the recognized shape class according to $t_j = \iota(j) \in G$.

\section{Results\label{sec:results}}


Like mentioned earlier, we solve the dictionary problem \eqref{eq:dict_learning} with the help of the \textit{DictionaryLearning} python class from the \textit{scikit-learn} library.
We set the number of atoms in the dictionary to $K=50$, the number of intervals to $N = 36$, and the number of non-zero coefficients in the representation vector $x$ to $T = 5$. The parameter $N=36$ indicates that we are scanning in $\approx10^\circ$ steps.

\subsection*{Recognition results}





\begin{figure}
	\centering
	\includegraphics[width = \columnwidth]{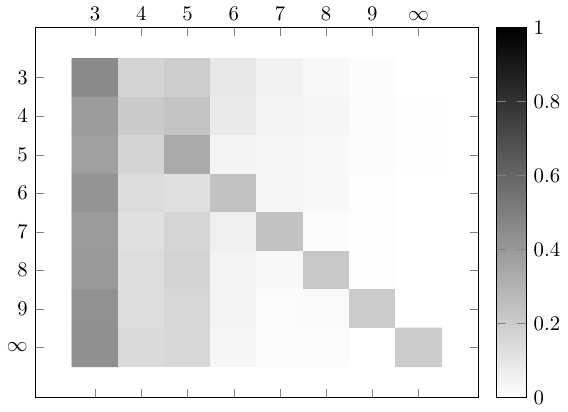}
	\caption{\label{fig:hrm_omp}
		The hit-rate-matrix when using the OMP algorithm as solver for \eqref{eq:dict_learning_subproblem_2}. Each row stands for the origin shape class, and the columns indicate the recognition results shape class.
		A strong diagonal line would indicate a perfect recognition, since the shape classes would be matched against them self.
		This figure indicates a poor recognition.
	}
\end{figure}

\begin{figure}
	\centering
	\includegraphics[width = \columnwidth]{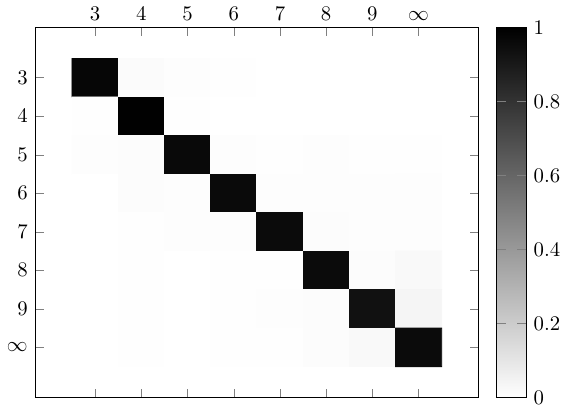}
	\caption{\label{fig:hrm_lars}
		The hit-rate-matrix when using the LARS algorithm as solver for \eqref{eq:dict_learning_subproblem_2}. Each row stands for the origin shape class, and the columns indicate the recognition results shape class.
		A strong diagonal line would indicate a perfect recognition, since the shape classes would be matched against them self.
		This figure indicates a very good recognition.
	}
\end{figure}

In this section, we show the results in form of the {\em hit-rate matrices} derived using one of the two algorithms to solve \eqref{eq:dict_learning_subproblem_2}. 
In the Figure \ref{fig:hrm_omp} the results using OMP are shown and in Figure \ref{fig:hrm_lars} presents the results using LARS.
The row represents the true shape class and the columns indicate the possible recognized shape classes.
Each cell in this $8 \times 8$-matrix is gray-value-color coded.
A value close to $1$ or $100\%$ is black, and white implies a recognition of $0\%$.
A good recognition would lead to a strong diagonal line, indicating that a shape class is matched against itself.
In contrast, strong side diagonal entries indicate a poor recognition.

In Figure \ref{fig:hrm_omp} we see an example of such poor recognition.
Strong diagonal elements are only slightly apparent for the shape classes with more corners, like hexagon and above.
However, the columns for triangle, square, and pentagon are mostly gray, indicating that, regardless of the original shape class, the recognized shape is likely to be one of these three.

The {\em hit-rate matrix} in Figure \ref{fig:hrm_lars} is an example of a very good recognition.
The diagonal elements are highly visible and almost pure black,
indicating, that the shape classes get matched to themselves with a probability of almost $100\%$.
For the nonagon shape class, we might notice a nearly unnoticeable gray entry for the circle shape class.

These results indicated that the atoms produced by the LARS method are more distinguished from each other.
Therefore, they produce a better (for the recognition task) overcomplete dictionary.

\section{Summary and Conclusion\label{sec:conclusion}}
In this paper, we presented a first approach using dictionary learning for shape recognition.
Specifically, we discussed the set-up of shape recognition and its evaluation,
which is taylored to the dictionary learning approach, and 
we compared the OMP and LARS algorithm in terms of resulting 
recognition capability. As a result, the LARS algorithm appears to be by far 
the better option for this task, giving nearly perfect results.

We conclude that SDL is potentially a suitable method 
for shape recognition
tasks, provided the set-up is properly done. 
Furthermore, to apply the SDL framework, one should
compare several methods and underlying parameters to ensure
its efficient use.


\bibliographystyle{IEEEtran}
\bibliography{references}

\end{document}